\documentclass[letterpaper, 10 pt, conference]{ieeeconf} 
\IEEEoverridecommandlockouts                              

\usepackage{cite}
\usepackage{amsmath,amssymb,amsfonts}
\usepackage{algorithm}
\usepackage[noend]{algpseudocode}

\usepackage{graphicx}
\usepackage{textcomp}
\usepackage{xcolor}
\usepackage{lipsum} 
\usepackage{hyperref} 
\usepackage{amsmath}
\usepackage{dsfont}
\usepackage{hyperref} 
\usepackage{cases}
\def\BibTeX{{\rm B\kern-.05em{\sc i\kern-.025em b}\kern-.08em
    T\kern-.1667em\lower.7ex\hbox{E}\kern-.125emX}}
\usepackage{caption}
\usepackage{subcaption}
\usepackage{pifont}
\usepackage{makecell}

\title{\LARGE \bf
End-to-End Learning of Deep Visuomotor Policy for Needle Picking
\\
\thanks{This work was supported in part by the Chow Yuk Ho Technology Centre of Innovative Medicine,
The Chinese University of Hong Kong, in part by the Multiscale Medical
Robotics Centre, AIR@InnoHK, and in part by the Research Grants Council
(RGC) of Hong Kong under Grants 14209118, 14209719, and 14211320. (Corresponding author: Xiangyu Chu)
}
\thanks{$^{1}$Hongbin Lin, Bin Li, Xiangyu Chu, Yunhui Liu and  Kwok Wai Samuel Au are with Department of Mechanical and Automation Engineering, The Chinese University of Hong Kong, Hong Kong. {\tt\small \{hongbinlin,binli\}@link.cuhk.edu.hk;  \{xiangyuchu,yhliu,samuelau\}@cuhk.edu.hk}}

\thanks{$^{2}$Qi Dou is with Department of Computer Science and Engineering, The Chinese University of Hong Kong, Hong Kong.
{\tt\small qdou@cse.cuhk.edu.hk}} %
}

\author{Hongbin Lin$^{1}$, Bin Li$^{1}$, Xiangyu Chu$^{1}$, Qi Dou$^{2}$, Yunhui Liu$^{1}$ and Kwok Wai Samuel Au$^{1}$
}

\begin{document}

\maketitle
\thispagestyle{empty}
\pagestyle{empty}

\begin{abstract}

Needle picking is a challenging manipulation task in robot-assisted surgery due to the characteristics of small slender shapes of needles, needles' variations in shapes and sizes, and demands for millimeter-level control. Prior works, heavily relying on the prior of needles (e.g., geometric models), are hard to scale to unseen needles' variations. In this paper, we present the first end-to-end learning method to train deep visuomotor policy for needle picking. Concretely, we propose \textit{DreamerfD} to maximally leverage demonstrations to improve the learning efficiency of a state-of-the-art model-based reinforcement learning method, DreamerV2; Since Variational Auto-Encoder (VAE) in DreamerV2 is difficult to scale to high-resolution images, we propose \textit{Dynamic Spotlight Adaptation} to represent control-related visual signals in a low-resolution image space; \textit{Virtual Clutch} is also proposed to reduce performance degradation due to significant error between prior and posterior encoded states at the beginning of a rollout. We conducted extensive experiments in simulation to evaluate the performance, robustness, in-domain variation adaptation, and effectiveness of individual components of our method. Our method, trained by 8k demonstration timesteps and 140k online policy timesteps, can achieve a remarkable success rate of 80$\%$. Furthermore, our method effectively demonstrated its superiority in generalization to unseen in-domain variations including needle variations and image disturbance, highlighting its robustness and versatility. Codes and videos are available at \href{https://sites.google.com/view/DreamerfD}{https://sites.google.com/view/DreamerfD}.

\end{abstract}

\section{Introduction}

\begin{figure*}[!tbp]
  \centering
  \includegraphics[width=1.0\hsize]{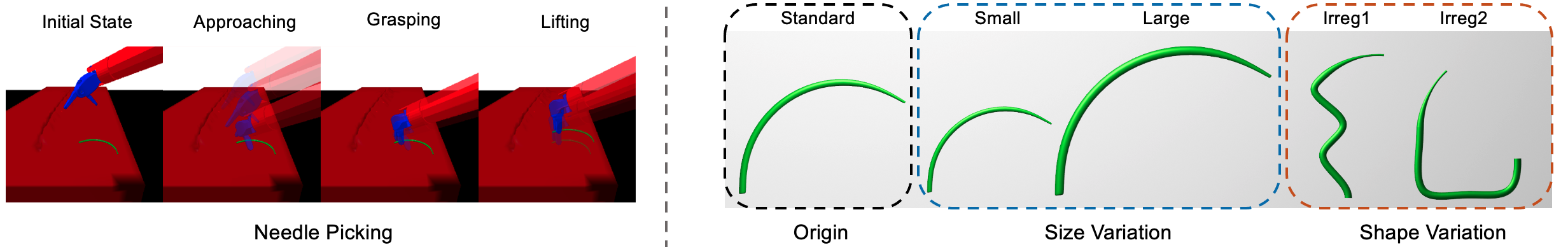}
  \caption{Canonical stages of needle picking (left) in RAS, including initial state, approaching, grasping, and lifting. In-domain variations in needle picking (right). We consider two instances for size variations and another two instances for shape variations.}
  \vspace{-0.45cm}
\label{fig:intro}
\end{figure*}

Needle picking is a repetitive and time-consuming task during surgery where suturing needles are required to be picked up by suturing tools (e.g., instruments) before stitching. Automating such a task in robot-assisted surgery (RAS) can greatly relieve surgeons' workload. Although algorithms have been proposed to solve the needle-picking tasks \cite{liu2015optimal,liu2016needle,d2018automated,sundaresan2019automated,chiu2021bimanual,xu2021surrol,wilcox2022learning}, the proposed algorithms entail pose information of key objects (e.g., needles and grippers of suturing tools). A few works studied image-based visual servo methods where projected points of the key objects were tracked for control without the need for pose estimation in needle-insertion \cite{zhong2019dual} and cutting tasks\cite{han2020vision}. However, their methods require designing hand-crafted features on the key objects. Neither the pose-estimation nor the feature-tracking methods are well adapted to unseen in-domain variations (e.g., shape or size variations of needles shown in Fig. \ref{fig:intro}) in needle-picking tasks due to the necessity of priors (e.g., geometric models of the key objects) and engineering efforts. 

Remarkable achievements in vision-based robotic tasks, including dexterous manipulation\cite{andrychowicz2020learning} and robot grasping\cite{kalashnikov2018scalable}, were obtained by model-free deep reinforcement learning (DRL). They learned deep visuomotor policies, which model the mapping from visual observations to robotic commands with deep neural networks and showed extraordinary abilities in adapting task variations and self-supervised learning. However, they required weeks of training on multiple real robots and distributed computation on multiple high-end GPUs, which prevented most practitioners from applying their methods due to the high cost of robotic maintenance and lack of robotic and computational resources.
\begin{figure}[!tbp]

  \centering
  \includegraphics[width=1.0\hsize]{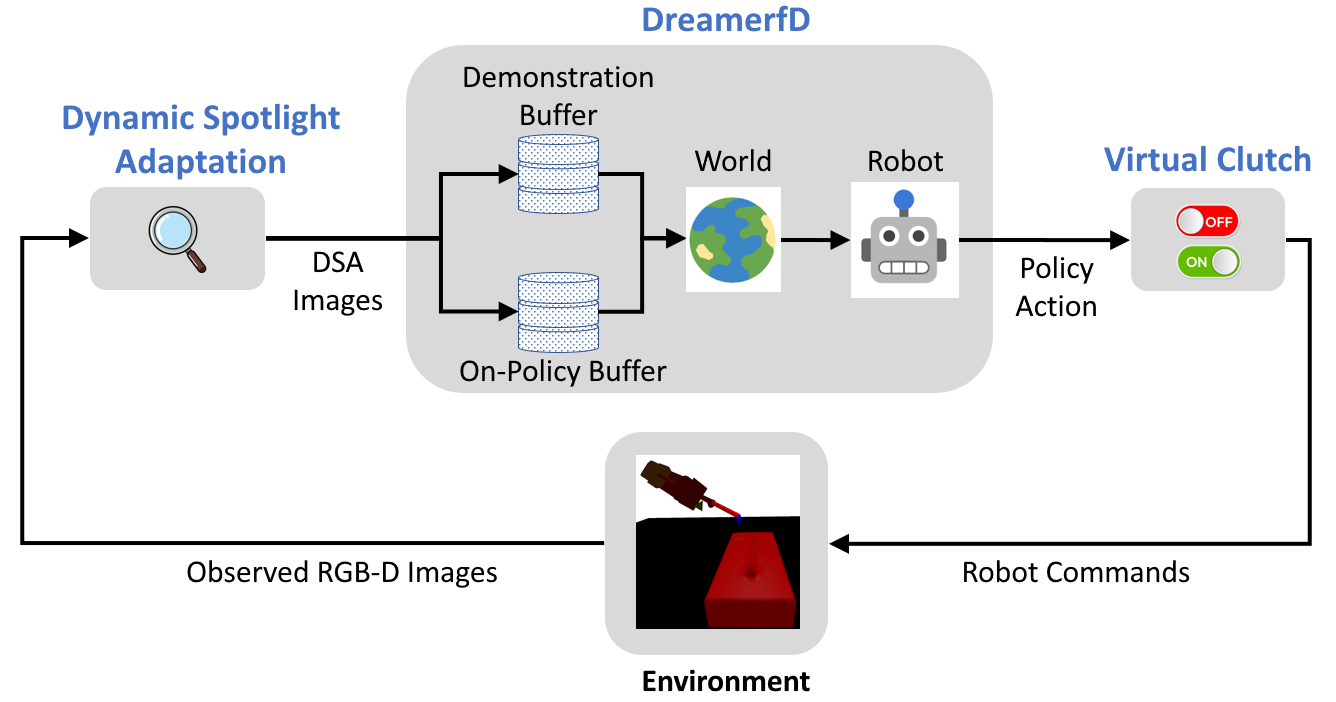}
  \caption{An overview of our method for end-to-end learning of deep visuomotor policy for needle picking: The visual inputs of visuomotor policy are obtained by pre-processing observed RGB-D images with our Dynamic Spotlight Adaptation; Robot commands are controlled by our Virtual Clutch with policy action inputs; Our learning method DreamerfD, which leverage demonstrations to maximally improve the learning of both world models and the policy\cite{hafner2020mastering}, learns our deep visuomotor policy unsupervisedly.}
  \vspace{-0.45cm}
\label{fig:method_overview}
\end{figure}

Model-based DRL, on the other hand, demonstrates astounding data efficiency and low computational cost while achieving competitive performances compared with the model-free DRL methods\cite{moerland2023model}. Recent advances in \textit{World Model}\cite{ha2018world,hafner2019planet}, which can predict action-conditioned future outcomes (e.g., visual observations, rewards), show state-of-the-art (SOTA) performances for end-to-end learning in video games \cite{hafner2019dream,hafner2020mastering} and robotic tasks\cite{wu2022daydreamer}. However, geometric characteristics of needles (e.g., millimeter-level size and slenderness) pose significant challenges to both vision and control considering deploying world-model-based DRL methods to needle-picking tasks.


In this work, we are interested in endowing robots with the ability to adapt in-domain variations for needle picking. To this end, a deep visuomotor policy was learned in a self-supervised manner guided by a small set of demonstration trajectories (See Fig. \ref{fig:method_overview}). We trained on a laptop with 8G RTX3070 GPU for 3 days, which was research-friendly due to low computation cost. To the best of our knowledge, we are 1) the first to deploy end-to-end learning of deep visuomotor policy for needle picking and 2) the first to investigate model-based DRL for learning deep visuomotor policy in surgical autonomy. Our main  contributions are:
\begin{enumerate}
    \item A general formulation for end-to-end learning of deep visuomotor policy in needle picking;
    \item A data-efficient model-based DRL framework, DreamerfD, that integrates a SOTA model-based RL framework, DreamerV2, with demonstrations; 
    
    \item Novel techniques, \textit{Dynamic Spotlight Adaptation} and \textit{Virtual Clutch}, to largely mitigate the issues of low-resolution image input and significant error between prior and posterior encoded states in DreamerV2, respectively;
    
    \item Systematic evaluation for the efficiency of our method, showing our performance, in-domain variation adaptation, robustness and effectiveness of individual components. We demonstrate our learned deep visuomotor policy can adapt to the needle's variations in shape and size.

\end{enumerate}

\section{Related Works}

\subsection{Combination of DRL and Demonstrations}
Popular choices of model-free DRL methods, such as PPO\cite{schulman2017proximal} and SAC\cite{haarnoja2018soft}, require a large amount of experience to solve general RL problems. Their sampling inefficiency is largely alleviated by model-based DRL (e.g. PILCO methods\cite{deisenroth2011pilco,gal2016improving}) and model-based planning (e.g. PETS\cite{chua2018deep}). When demonstrations are applicable, DRL can be combined with demonstrations \cite{vecerik2017leveraging,nair2018overcoming,goecks2019integrating} to accelerate learning efficiency and is proven to be effective in sparse-reward settings. However, scaling the aforementioned methods to RL scenarios with high dimensional observation is challenging.

Recent advances in modeling the dynamics of the environment show promise in scaling DRL to the setting of high-dimensional observation. Visuomotor policies can be optimized by interacting learned dynamic models in image space (Visual Forsight\cite{ebert2018visual}) or latent space (world-model-based DRL\cite{hafner2019dream,hafner2020mastering}). Alternatively, the learned pixel-level dynamic model can be used to generate optimal trajectories using planning methods\cite{hafner2019planet,koul2020dream,schrittwieser2020mastering}. However, how to combine world-model-based DRL with demonstrations efficiently remains unclear. In contrast, we focus on how to maximally leverage demonstrations to improve the learning of world-model-based DRL. 

\subsection{Needle Picking}

Researchers have studied needle picking extensively: Liu et al. investigated the optimal grasping poses of a needle\cite{liu2015optimal,liu2016needle}; Ettorre et al. first achieved needle picking without the assistance of angular positioners\cite{d2018automated}; Sundaraesan et al. further achieved needle picking in settings of needle occlusion\cite{sundaresan2019automated}. Another surgical task related to needle picking is needle re-grasping, where a robotic arm is required to hand over its grasped needle to another robotic arm: Chiu et al.\cite{chiu2021bimanual} applied BC-integrated DDPG \cite{nair2018overcoming} to plan viable trajectories in decision time; Wilcox et al.\cite{wilcox2022learning} achieved needle re-grasping in the setting of needle occlusion. Recently, Researchers \cite{xu2021surrol} demonstrated a strong ability of generalization using model-free DRL methods, where control policies were learned in simulation and then transferred the policy to real robots for multiple surgical tasks including needle picking, needle re-grasping, etc. However, the aforementioned methods, requiring either tracking needle poses or features, heavily relied on the prior of needles (e.g., the geometric model of needles), limiting the adaptation ability to unseen in-domain variations (e.g., variations of needle shape and size). The closest work was from Scheikl et, al.\cite{scheikl2022sim}, who trained deep visuomotor policy with model-free DRL in simulation and transferred it to real robots with an Unpaired Image-To-Image translation model for tissue retraction. Nevertheless, the dense reward they used entailed tracking positions of key points (e.g., goal points on the tissue) and extensive reward engineering, which hardly improve the scalability in in-domain variations.


\section{Problem Description}
\label{sec:problem description}
We focus on solving a set of needle-picking tasks for RAS, where a needle on a plane is required to be picked up by a robotic arm. The initial poses of the needle and the gripper are random. RGB-D images $I_t$ from a monocular camera are observed at time $t$.  Discrete commands are used to servoing the robotic arm incrementally. Specifically, the robotic gripper can be driven translationally (along X, Y, and Z axes for 2 mm), and rotationally (along the normal axis of the plane for 10 Deg) in both positive and negative direction w.r.t. a fixed world frame; The jaw of the gripper can be opened and closed by a toggling command; All commands are decoupled, resulting in discrete commands $a_t$ in a 9-element discrete set $A$ at time $t$. The tasks of needle picking have a finite task horizon $T=100$. The task is successful if the needle is grasped and lifted to 0.06 mm above the plane by the gripper. The goal of our work is to develop a control policy that maps from historical observed images to control actions. We assume that 1) the robotic arm will not exceed its joint limits, and 2) either the needle or the gripper can be partially but not fully occluded.

\section{Preliminaries}
\subsection{Problem Formulation}
We model the needle-picking task as a discrete-time partially observable Markov Decision Process (POMDP) with discrete action space, which can be formally described as a 7-tuple $(S,A,T,R,\Omega,O,\gamma)$. The elements in the 7-tuple are defined as follows: $S$ is a set of partially observable states; $R(s, a): S\times A \to \mathbb{R}$ is a reward function; $T$ is a set of conditional transition probabilities between states; $O$ is a set of conditional observation probabilities; $\gamma \in [0,1]$ is the discount factor; $\Omega$ is the observation and $A$ is a set of action defined in Section \ref{sec:problem description}. The goal is to learn a control policy $\pi$ that maximizes its expected future discounted reward
$\mathbb{E}_{\pi}[ \sum_{i=t}^{T} \gamma^{i-t}r_{i}]$, where $r_i$ is the reward at time $i$ and $t$ is the current time.

\subsection{DreamerV2}
DreamerV2\cite{hafner2020mastering}, an advanced model-based RL method, consists of 2 components: world models, targeting to model the dynamics of environments with POMDP formulation, and a control policy, targeting to learn from merely simulated trajectories in world models and generate rollout trajectories for the learning of world models.


For world models, observation $x_t$, including images and scalar signals, is encoded to stochastic latent state $z_t$ through Variational Auto-Encoder (VAE)\cite{kingma2013auto}. Then, the sequences of the latent states are predicted by Recurrent State-Space Model (RSSM)\cite{hafner2019planet}, a sequence model with a deterministic recurrent state $h_t$. Finally, the reward $r_t$, the discount factor $\lambda_t$, and reconstructive observation $\hat{x}_t$ are predicted based on the model state, which is formed by the concatenation of these states as $s_t=[h_t, z_t]$. In summary, the world model can be described as follows:
\begin{equation}
\begin{array}{ll}
\text{RSSM} & \left\{ \begin{array}{l}
\text{Recurrent model:} \quad h_t = f_{\phi}(h_{t-1},z_{t-1},a_{t-1})\\
\text{Representation model:} \quad z_t \sim q_{\phi}(z_t | h_t, x_t)\\
\text{Transition predictor:} \quad  \hat{z}_t \sim p_{\phi}(z_t | h_t)\\
\end{array}\right.\\
&\quad\begin{array}{l}
 \text{Image predictor:} \quad \hat{x}_t \sim p_{\phi}(\hat{x_t} | h_t, z_t)\\
\text{Reward predictor:} \quad \hat{r}_t \sim p_{\phi}(\hat{r}_t | h_t, z_t) \\
\text{Discount predictor:} \quad \hat{\lambda}_t \sim p_{\phi}(\hat{\lambda}_t | h_t, z_t),
\end{array}
\end{array}
\end{equation}
where $p$ and $q$ denote prior and posterior distributions, respectively; $\phi$ is the parameter of world models, which can be learned by optimizing the world model loss as:
\begin{equation}
\label{equ:wolrd_model_standard_loss}
\begin{aligned}
     \mathcal{L}_{\phi}= &\mathbb{E}_{q_\phi(z_{1:T}|a_{1:T},x_{1:T})}\big[ \sum^{T}_{t=1}-\ln p_{\phi}(x_t r_t \gamma_t | s_t) 
    \\&+\beta_{kl} KL[q_{\phi}(z_t|s_t)||p_{\phi}(z_t|h_t)]\big],
\end{aligned}
\end{equation}
where $KL$ and $\beta_{kl}$ are the KL divergence loss and its scaling weight, respectively.

The control policy is learned by the actor-critic mechanism, where an actor  $p_{\psi}(\hat{a}_t | \hat{z}_t)$ and a critic $v_{\xi}(\hat{z}_t)$ are used to predict the action and the value, respectively.
The critic network, predicting the $\lambda$-return $V_t^{\lambda}$, is learned by optimizing a mean square loss as:
\begin{equation}
\label{equ:critic_standard_loss}
    \mathcal{L}_{\xi}= \mathbb{E}_{p_{\phi},p_{\psi}}\frac{1}{2}[\sum^{H-1}_{t=1}(v_{\xi}(\hat{z}_t)-sg(V_t^{\lambda}))^{2}], 
\end{equation}
where $sg(\cdot)$ denotes the stop gradient function. The actor is learned by the Reinforce algorithm\cite{williams1992simple}, where the policy loss is designed as
\begin{equation}
\label{equ:actor_standard_loss}
\begin{aligned}
    \mathcal{L}_{\psi}= \mathbb{E}_{p_{\phi},p_{\psi}}[\sum^{H-1}_{t=1}&-\beta_{r}\underbrace{\ln p_{\psi}(\hat{a}_t | \hat{z}_t)sg(V^{\lambda}_t-v_{\xi}(\hat{z}_t))}_{reinforce}\\
    &-\underbrace{\beta_{e} H[a_t|\hat{z}_t]}_{entropy}], 
\end{aligned}
\end{equation}
where $H$ is the loss of an entropy regularizer to incent exploration; $\beta_{r}$ and $\beta_{e}$ are scaling weights for the Reinforce algorithm and the entropy regularization, respectively.

The overall loss of DreamerV2 can be formulated as
\begin{equation}
\label{equ:overall_dreamevv2}
\mathcal{L}_{_{\text{DV2}}} = \mathbb{E}_{(x_{0:T},r_{1:T},\lambda_{1:T},a_{1:T})\sim D}[\mathcal{L}_{\phi}] + \mathbb{E}_{x_i\sim D}[\mathcal{L}_{\xi}+\mathcal{L}_{\psi}],
\end{equation}
where transitions are sampled from policy experiences $D$ with a replay buffer\cite{vecerik2017leveraging}.
Details of Dreamerv2 can be found in \cite{hafner2020mastering}.


\section{DreamerfD: Integrating World Models and demonstrations}
\label{sec:DreamerfD}
We present \textbf{Dreamer} \textbf{f}rom \textbf{D}emonstrations (\textbf{DreamerfD}) to maximally enhance the learning efficiency of DreamerV2 with demonstrations for POMDP with sparse delayed rewards and high-dimensional observation. 

First, a set of suboptimal demonstration trajectories $D_E$, which normally have better returns compared to rollout trajectories of DreamerV2 at the early learning stage, is gathered by either human demonstration or hand-engineered programs. Then, the demonstration trajectories are added to an additional replay buffer and keep all transitions during training. The demonstration replay buffer samples transitions of demonstrations to train the world models and the control policy, where the training loss can be defined by reformulating (\ref{equ:overall_dreamevv2}) as
\begin{equation}
\label{equ:exposed_loss}
\mathcal{L_{\text{E}}} = \mathbb{E}_{(x_{0:T},r_{1:T},\lambda_{1:T},a_{1:T})\sim D_E}[\mathcal{L}_{\phi}] + \mathbb{E}_{x_i\sim D_E}[\mathcal{L}_{\xi}+\mathcal{L}_{\psi}];
\end{equation}
Besides, we apply Behavior Cloning (BC) to guide the learning of the policy with demonstrations, where a BC loss is used to train the actor to regress actions in $D_E$ as  
\begin{equation}
    \mathcal{L}_{\text{BC}}= \mathbb{E}_{(x_{0:T},a_{1:T})  \sim D_{E}}\big[\beta_{bc}\underbrace{\mathbb{E}_{p_{\phi},p_{\psi}}\big[ \sum^{T}_{t=1}
    -\ln p_{\psi}(a_t|s_t)\big]}_{BC}\big],
\end{equation}
where $\beta_{bc}$ is a scaling weight for BC; Finally, the overall loss for DreamerfD can be written as:
\begin{equation}
\mathcal{L}=\mathcal{L}_{\text{DV2}} + \mathcal{L}_{\text{E}} +\mathcal{L}_{\text{BC}};
\end{equation}
Note that we train multiple objectives simultaneously without multi-stage training. In summary, we modified the DreamerV2 algorithm as follows:
\begin{itemize}
    \item Transitions of demonstrations are added to the additional replay buffer;
    \item An additional training loss, sampling from demonstrations instead of learned policy experiences, is used to train both the world models and the control policy;
    \item An additional BC loss, sampling from the demonstration replay buffer, is used to guide the policy learning.
\end{itemize}

 \begin{figure}[!tbp]
  \centering
  \includegraphics[width=1.0\hsize]{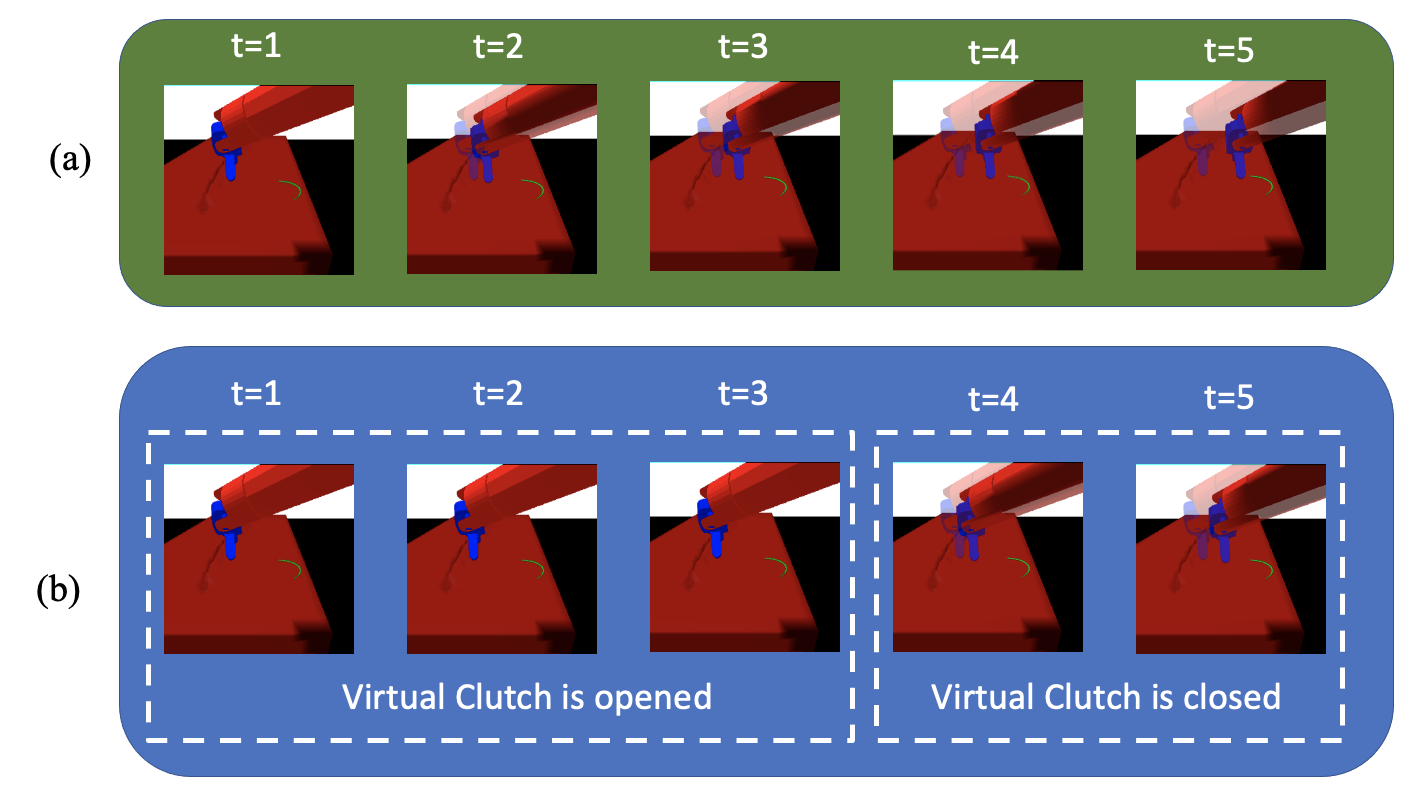}
  \caption{Schematic illustration for Virtual Clutch. (a) shows the first 5 timesteps of a rollout when Virtual Clutch is not applied to our visuomotor controller. (b) shows the corresponding 5 timesteps after applying Virtual Clutch ($H_{clutch}=4$).}.
  \vspace{-0.45cm}
\label{fig:VC}
\end{figure}
\section{End-to-end Deep Visuomotor Control for Needle picking}
In this section, we will show how to deploy our end-to-end learning method to needle-picking tasks. First, we will introduce task-level states determined by a Finite State Machine for needle picking. Then, we will elaborate on two proposed techniques, \textit{Dynamic Spotlight Adaptation} and \textit{Virtual Clutch}, that dramatically improve image representation and mitigate the issue of significant error between prior and posterior encoded states, respectively.

\subsection{Task-level Finite State Machine}
\label{sec:task-level state}
We design a Finite State Machine (FSM) that directly determines task termination and reward assignment. The task-level states in FSM are defined as:
\begin{itemize}
    \item Successed Terminated State: The needle is successfully grasped. In this state, the reward is $1$ and the task is terminated.
    \item Failed Terminated State: The task failed due to the limit of task horizon $T$. In such a case, the reward is $-0.1$ and the task is terminated.
    \item Failed Non-terminated State: The task failed due to exceeding the workspace. In such a case, the reward is $-0.01$ and the task is not terminated. In addition, the actor action will be replaced by $a_{idle}$ for the control output.
    \item In-progress State. The task is performed in progress. The reward is $-0.001$ and the task is not terminated.
\end{itemize}

\subsection{Virtual Clutch for Deep Visuomotor Control}
\label{sec:virtual_clutch}
There is a significant error between prior and posterior state $\hat{z}_t$ and $z_t$ at the beginning of a rollout due to poor initial prior guess (e.g., a zero matrice in \cite{hafner2019dream}) and slow convergence on the error. We empirically observed that the controller of DreamerV2 performs poorly for the first few timesteps due to large errors, leading to catastrophic results, (See the results in our ablation study in the later Sec. \ref{sec:ablation_study}). 

\begin{figure*}[!tbp]
  \centering
  \includegraphics[width=1\hsize]{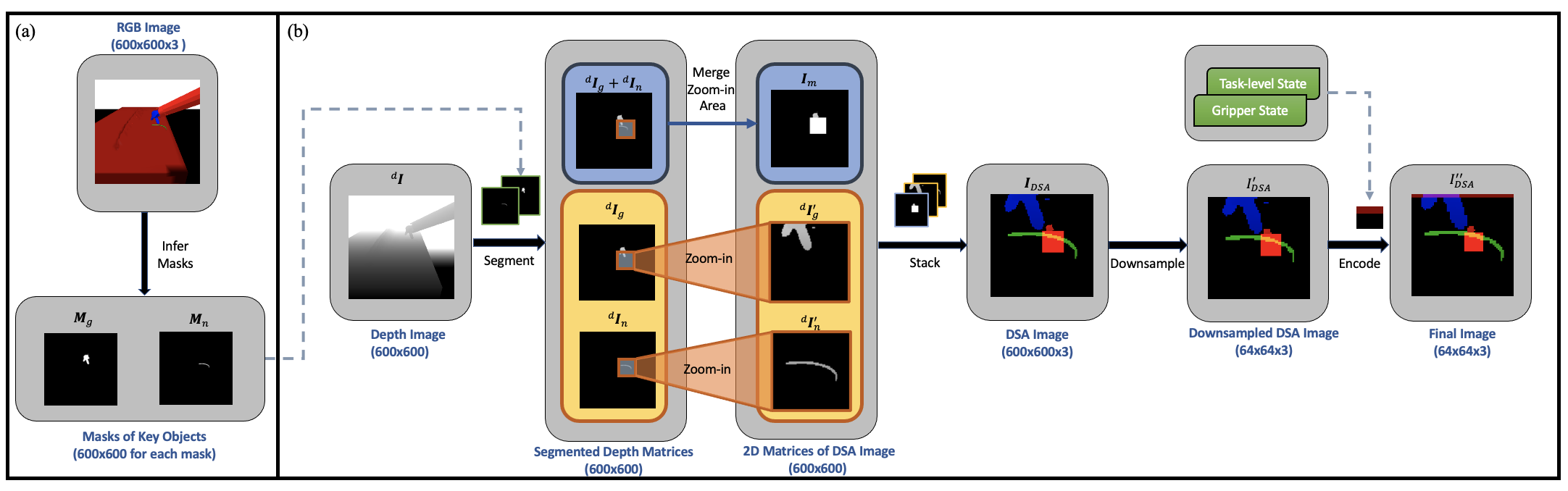}
  \caption{Pipeline of Dynamic Spotlight Adaptation (DSA) for needle picking. (a) shows that segmented masks of key objects are inferred by a color-based scripted program based on original RGB images. In (b), depth matrices are segmented by the inferred masks. 2D matrices of DSA images can be obtained by zooming in needle-centric areas in the segmented depth matrics and merging the zoom-in area with out-of-zoom depth pixels. We stack the 2D matrices to a 3D DSA image, followed by downsampling. The final image is obtained by merging the resultant downsampled image with the image encoding of task-level states and gripper states.}
  \vspace{-0.45cm}
  \label{fig:DSA}
\end{figure*}
We propose a simple technique, \textit{Virtual Clutch}, to solve this issue. Our inspiration is from the vehicle clutch design where vehicles can only be driven when the clutch is closed. Similarly, the control output of our controller $a_t^c$ is determined by a timestep-dependent clutch as
\begin{equation}
    a_t^c = \begin{cases} a_t, & t>=H_{clutch}\\
    a_{idle}, & t<H_{clutch},
    \end{cases}
\end{equation}
where $a_{idle}$ is an idle action command that keeps the joint positions of the robot arm unchanged, $H_{clutch}$ is a non-negative constant determining the timestep that starts to close the virtual clutch, and $a_t$ is the action from the learned agent policy. Fig. \ref{fig:VC} shows the illustration for Virtual Clutch.

\subsection{Dynamic Spotlight Adaptation for Image Representation}
\label{sec:dsa}
Learning the visual component of DreamerV2 for needle picking is challenging: First, VAE is notoriously hard to scale to high-dimensional images\cite{kingma2013auto}, and thus origin RGB images were down-sampling to 64x64x3 low-dimension RGB images due to limitations of the VAE capacity in DreamerV2\cite{hafner2020mastering}, leading to degradation of geometric and pose-related visual information, especially for needle-picking tasks where the needles are tiny and slender (See the degradation results in our ablation study in the later Sec. \ref{sec:ablation_study}); (b) Visual noises (e.g., the background that is irrelevant to our control), taking majority proportion in origin images, especially for RAS, are encoded by VAE using RGB-D visual representation, which is both data-inefficient and computationally expensive.  

We propose \textit{Dynamic Spotlight Adaptation} (DSA) for image representation, which maximally represents control-related visual information on 64x64x3 low-dimensional image space. We are inspired by the Mini-Map mechanism in computer games of multiplayer online battle arena (MOBA): local information can be observed by zooming in the agent-centered area while the global information of the environment (e.g., other out-of-frame agents, the location of the zoom-in area) can be observed by a mini-map. Similarly, DSA is obtained by the following pipeline: First, we segment the region of interest of key objects (i.e., the needle and the gripper), where the binary image masks, $\boldsymbol{M}_{n}$ and $\boldsymbol{M}_{g}$, are inferred by a simple script of color segmentation for the needle and the gripper, respectively. Then, the depth-channel matrix of the original RGB-D image ${}^{d}\boldsymbol{I}$ is segmented by the inferred masks for both needle and gripper as
\begin{equation}
\label{equ:segment_matrix}
    {}^{d}\boldsymbol{I}_{g} = {}^{d}\boldsymbol{I} \odot \boldsymbol{M}_{g},  \quad   {}^{d}\boldsymbol{I}_{n} = {}^{d}\boldsymbol{I} \odot \boldsymbol{M}_{n},
\end{equation}
where $\odot$ is a dot product operator, and ${}^{d}\boldsymbol{I}_{g}$ and ${}^{d}\boldsymbol{I}_{n}$ are the corresponding segmented matrices for the gripper and needle, respectively. Next, the segmented matrices are processed by zoom-in (the segmented matrices are cropped with a bounding square box $\boldsymbol{B}$, followed by resizing the resultant matrices), which is defined as
\begin{equation}
    ^{d}\boldsymbol{I}^{'}_{g} = f_{rs}(f_{crop}({}^{d}\boldsymbol{I}_{g}, \boldsymbol{B})), \quad  {}^{d}\boldsymbol{I}^{'}_{n} = f_{rs}(f_{crop}({}^{d}\boldsymbol{I}_{n}, \boldsymbol{B})),
\end{equation}
where $f_{crop}$ and $f_{rz}$ are the image crop and resize function, respectively; $\boldsymbol{B}$ is the square zoom-in box which has the same center coordinates with the bounding box of the needle mask $\boldsymbol{M}_{n}$ and a side length $b\in [1,+\infty)$ times the maximum side length of the bounding box, considering leaving image margins; $^{d}\boldsymbol{I}^{'}_{g}$ and ${}^{d}\boldsymbol{I}^{'}_{n}$ are zoom-in depth matrices for the gripper and the needle, respectively. Besides, an additional matrice $\boldsymbol{I}_{m}$, which is a mixed information of both the zoom-in box and out-of-zoom pixels in the origin depth image, is formulated as
\begin{equation}
    \boldsymbol{I}_{m} = f_{rs}(f_{clip}(\boldsymbol{B} + {}^{d}\boldsymbol{I}_{g}+ {}^{d}\boldsymbol{I}_{n})),
\end{equation}
where $f_{clip}$ is a clip function that saturates the input value to the range $[0, 255]$ for the 8-bit unsigned integer. We stack these processed 2-D matrices to a 3-D matrice $\boldsymbol{I}_{DSA}$ as
\begin{equation}
    \boldsymbol{I}_{DSA} = \begin{bmatrix}\boldsymbol{I}_{m}&{}^{d}\boldsymbol{I}^{'}_{n}& ^{d}\boldsymbol{I}^{'}_{g}
    \end{bmatrix}.
\end{equation}
Finally, we obtain the final image $\boldsymbol{I}_{DSA}^{''}$ for DSA: $\boldsymbol{I}_{DSA}$ is downsampled to 64x64x3 image $\boldsymbol{I}_{DSA}^{'}$, and then we encode the scalar signals, including task-level states in Sec. \ref{sec:task-level state} and the toggling command of the gripper, to 3D image matrices using broadcasting (a common practice in \cite{schrittwieser2020mastering}), followed by adding the encoded matrices to the downsampled DSA image. The final DSA image can be obtained by
\begin{equation}
   \boldsymbol{I}_{DSA}^{'} = f_{rz}(\boldsymbol{I}_{DSA}),  \quad   \boldsymbol{I}_{DSA}^{''}= \boldsymbol{I}_{DSA}^{'} + \boldsymbol{I}_{e},
\end{equation}
where $\boldsymbol{I}_{DSA}^{'},\boldsymbol{I}_{DSA}^{''},\boldsymbol{I}_{e}\in\mathcal{R}^{64\times64\times3}$. Fig. \ref{fig:DSA} illustrates the pipeline of DSA in detail.

\section{Experiments}

Extensive experiments were carried out in the simulation, aiming to answer the following questions:
\begin{itemize}
  \item How is the performance of our method compared to the SOTA framework?
  \item Can our method adapt to unseen task in-domain variations?
  \item Is our method robust to unseen noises and disturbances?
  \item How effective are the individual components of our methods? 
\end{itemize}
We start with introducing the experiment setup of simulation, followed by elaborating on the studies of performance, in-domain variation adaptation, robustness, and ablation.

\subsection{Experiment Setup}
\begin{figure}[!tbp]
  \centering
  \includegraphics[width=0.8\hsize]{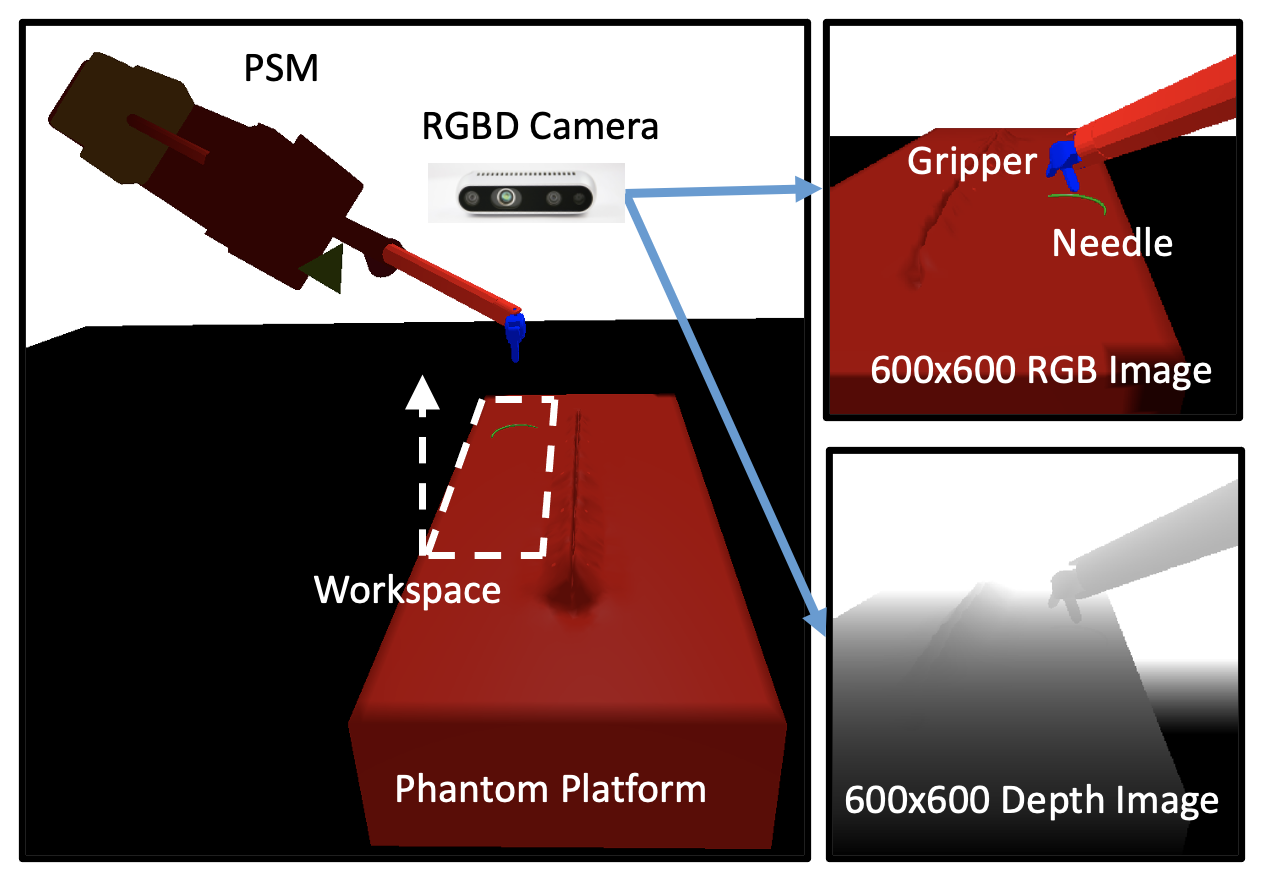}
  \caption{Experiment setup in the simulation of AccelNet Surgical Robotics Challenge\cite{munawar2022open}.}
  \vspace{-0.45cm}
  \label{fig:setup}
\end{figure}

We conducted experiments on the simulation platform of 2021-2022 AccelNet Surgical Robotics Challenge \cite{munawar2022open}, which provided high-fidelity simulation, standardized problem definitions, and evaluations for benchmarking autonomous suturing (See Fig. \ref{fig:setup}). We developed our simulation code based on our previous work\cite{hlin2022Accel}, which ranked first among all competitors. The dimension of the workspace was $40mm \times 60mm \times 30mm$; A third-person camera provided $600\times600\times4$ RGB-D images for visual feedback. Synthetic colors, (blue for the gripper, green for the needle, and red for the background), were used to reduce the difficulty of our color-based segmentation since segmentation was beyond the focus of this paper. The success rate was chosen as the key index of our evaluation, similar to that of \cite{wilcox2022learning}. Rollouts beyond our assumptions in the evaluation were neglected. For each training of our method and baselines, the maximum training timesteps $L$ was $140000$; $20$ evaluation rollouts were carried every $2000$ training timestep for the evaluation of the success rate; $8000$ timesteps of demonstrations, (approximately 400 rollouts), were pre-filled using a script developed in our previous work\cite{hlin2022Accel}; Each training of the experiment was run on a laptop with RTX 3070 GPU with $8$ Gigabite Memory for 3 days. Details of key hyperparameters can be found in Appendix \ref{sec:appendix_a}.

\subsection{Performance Study}
We compared our method with SOTA end-to-end learning framework, DreamerV2 \cite{hafner2020mastering}, in simulation. To keep the image observation space the same as ours for a fair comparison, the original RGB-D images were transformed into mixed images with 3 channels, where one channel was obtained by the weighted summation of the original red and blue channels, and the remaining two channels were the original blue and depth channels. The resultant mixed images were downsampled to $64\times64\times3$ images. Scalar signals, (i.e., task-level states and gripper states), were encoded by Multi-Layer Perceptron\cite{hafner2020mastering,wu2022daydreamer}. The remaining training settings for the baseline were kept consistent with our approach. 

Fig. \ref{fig:performance_study} shows the results of the performance study. Our method converged after $134$k training timesteps, reaching $80\%$ success rate in the evaluation. Compared to our method, the DreamerV2 baseline had not yet converged after $140$k training timesteps. Although we observed incremental improvement at around $20$k, $60$k, and $125$k training timesteps, the success rate was extremely low for most of the training timesteps. Overall, our method demonstrated higher data efficiency in learning visuomotor policy for needle picking.









\subsection{In-domain Variation Adaptation Study}
A wide variety of needles to study the ability to adapt in-domain variations were designed (See Fig. \ref{fig:intro}). The needle in AccelNet Surgical Robotics Challenge \cite{munawar2022open} served as the standard needle. For the size variation, a small needle and a large needle were obtained by scaling the standard needle by the factor of $0.75$ and $1.3$, respectively. For the shape variation, we designed two irregular needles the same size as the standard needle. Note that our visuomotor policy was only trained on the standard needle, and then transferred directly to the settings of unseen in-domain variations. 

\begin{figure}[!tbp]
     \centering
     \begin{subfigure}[b]{0.5\textwidth}
  \centering
  \includegraphics[width=0.8\hsize]{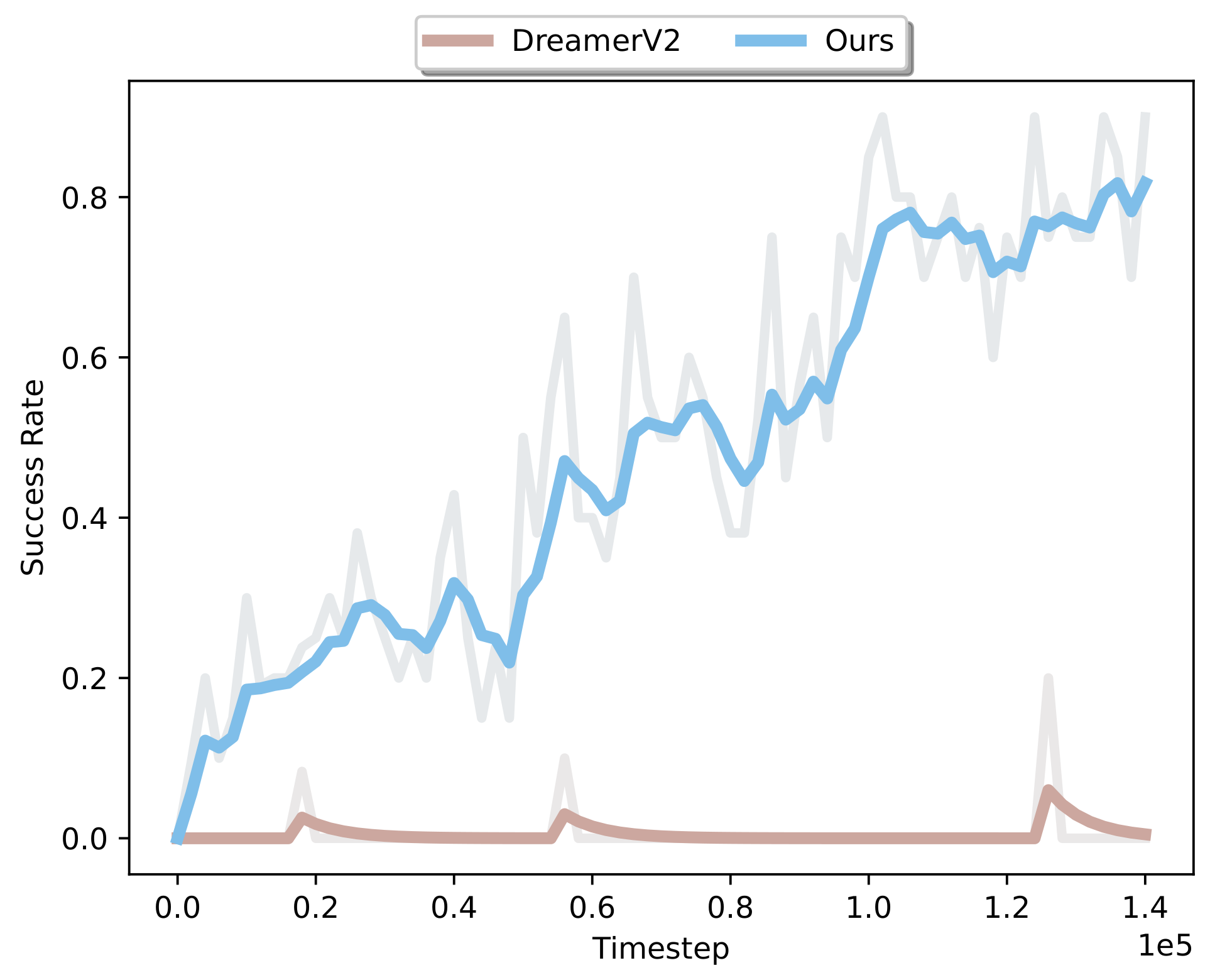}
  \caption{Performance study}
  \label{fig:performance_study}
     \end{subfigure}
     \hfill
     \begin{subfigure}[b]{0.48\textwidth}
         \centering
         \includegraphics[width=\textwidth]{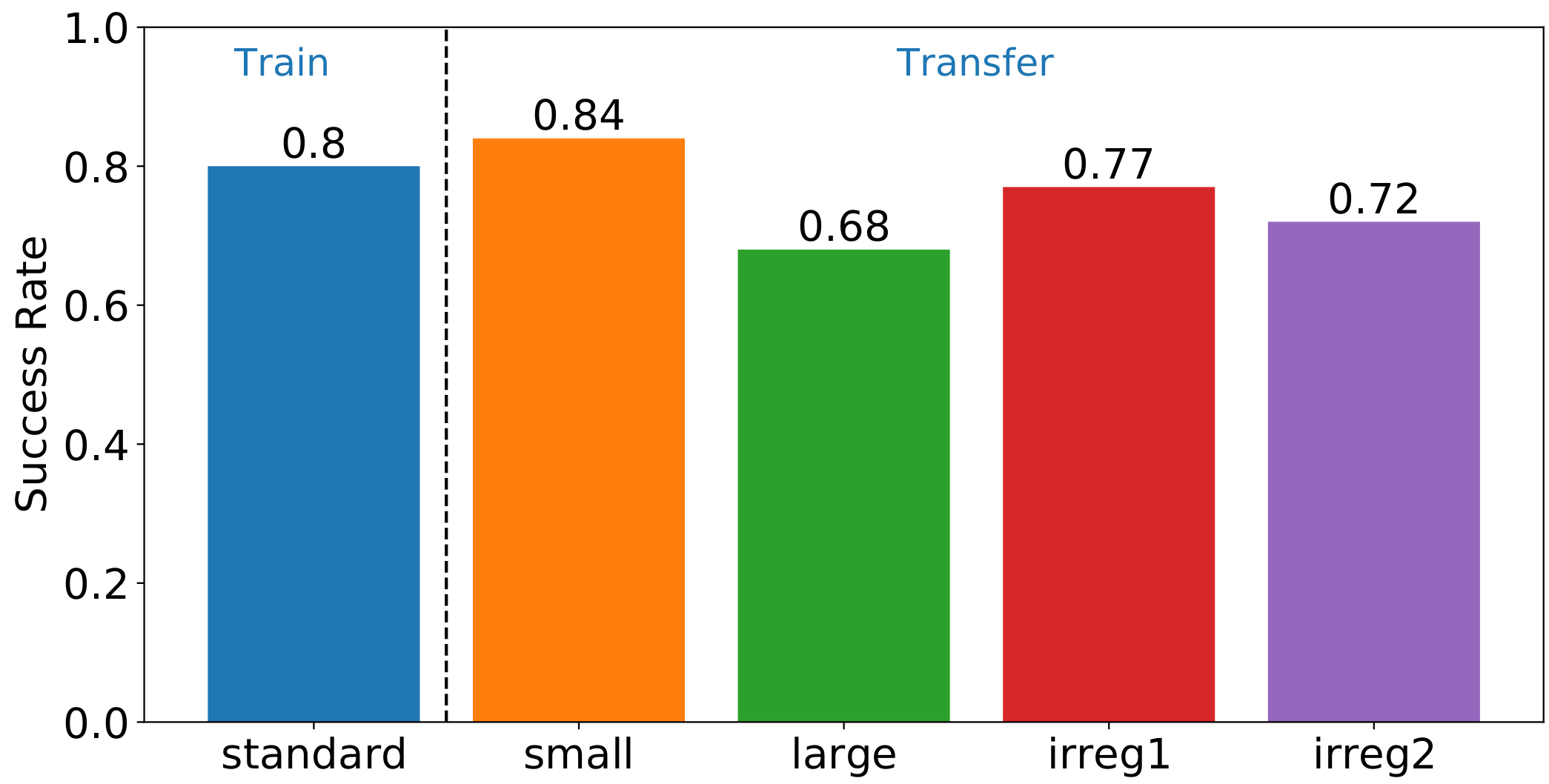}
         \caption{In-domain variation adaptation study}
         \label{fig:variation_performance}
     \end{subfigure}
     \caption{Evaluations in (a) performance study and (b) in-domain variation adaptation study. In (a), the evaluation curves of success rate for the training process were demonstrated; Curves are smoothed by Exponential Moving Window with a $0.7$ smoothing factor. In (b), our controller was first trained based on the standard needle and then transferred to unseen needles with variations (See needle instances in Fig. \ref{fig:intro}). For each needle, the success rate was evaluated with 100 rollouts.}

\end{figure}

Fig. \ref{fig:variation_performance} shows the performance of our trained policy applied to unseen needles. The standard needle served as the baseline. We observed that our model achieved a higher success rate of 84\% for grasping small needles because of its higher error tolerance, and achieved similar performance in grasping varied irregular shape needles with the same size compared to the baseline. Although the grasping larger size needle requires a more accurate claw pose, as thicker width reserves less space for opened claw with a certain width which means less error tolerance, our method can still obtain a success rate of up to $68\%$. The above promising results demonstrated our visuomotor policy can be efficiently adapted to unseen needle variants.

\subsection{Robustness Study}
\begin{figure}[!tbp]
     \centering
     \begin{subfigure}[b]{0.4\textwidth}
         \centering
         \includegraphics[width=\textwidth]{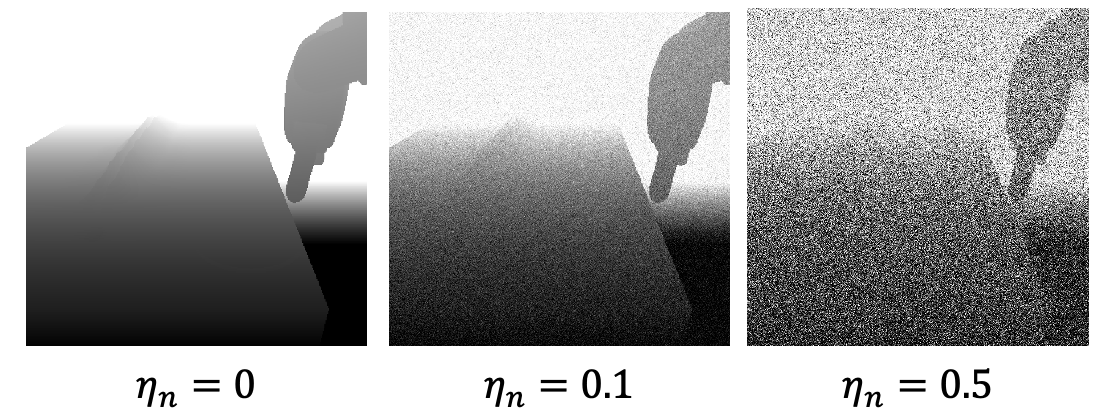}
         \caption{Depth images under different noise levels}
         \label{fig:robustness_illustration}
     \end{subfigure}
     \hfill
     \begin{subfigure}[b]{0.5\textwidth}
         \centering
         \includegraphics[width=0.8\textwidth]{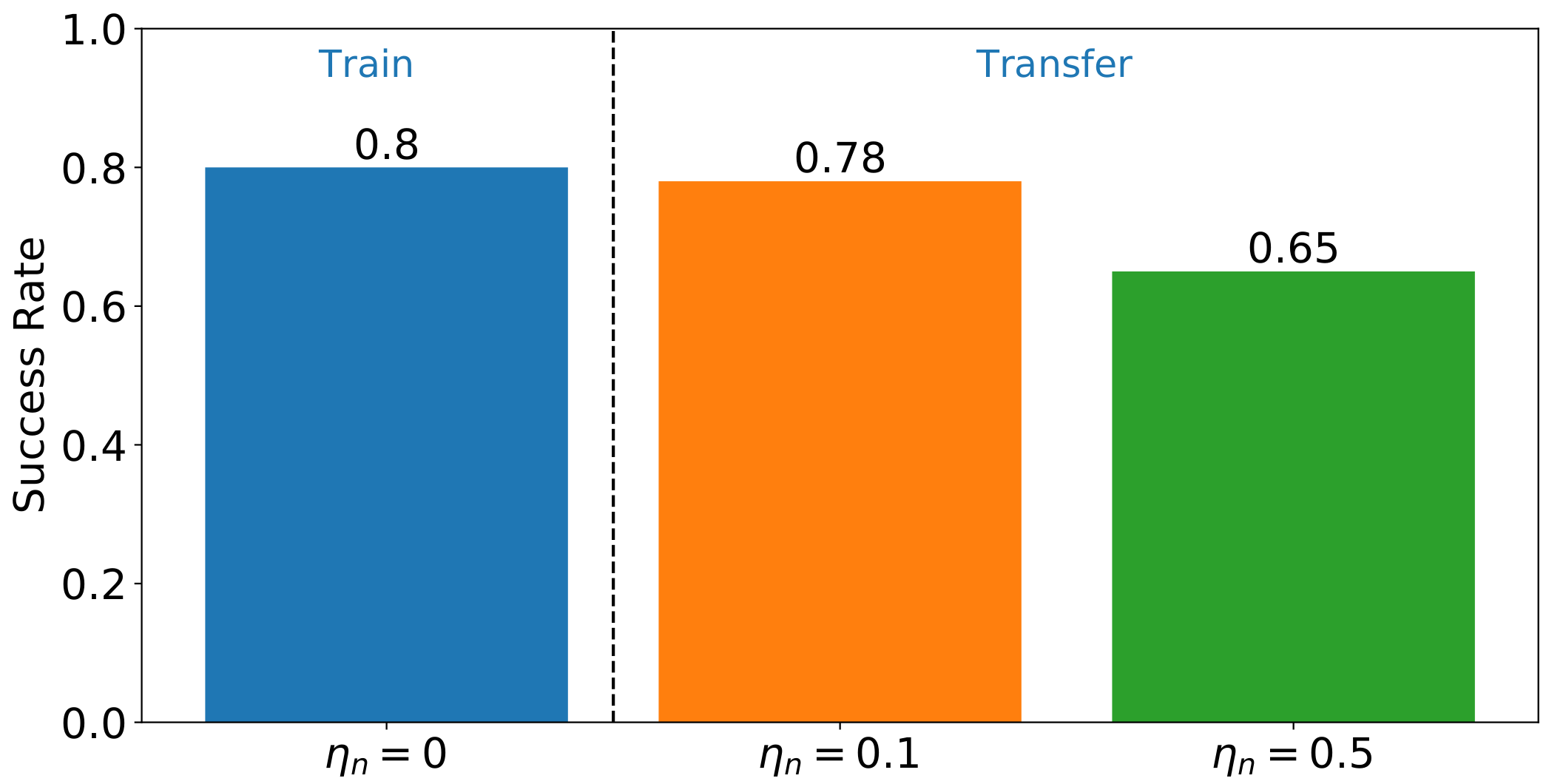}
         \caption{Performance in robustness study}
         \label{fig:robustness_performance}
     \end{subfigure}
     \caption{Robustness study. In (a), we show 3 different noise levels for depth images. In (b), our controller was first trained in the environment with noise-free depth images ($\eta_{n}=0$) and then transfer to environments with noisy depth images ($\eta_{n}=0.1$ and  $\eta_{n}=0.5$). For each baseline, the success rate was evaluated with 100 rollouts.}
\end{figure}

We evaluated the robustness of our method quantitatively and qualitatively. For the quantitative evaluation of robustness, we transferred our trained policy to unseen environments with two different noise levels of depth images. In particular, gaussian noises which samples from the unit Gaussian distribution scaled by a factor of $255\cdot\eta_{n}$ (See Fig. \ref{fig:robustness_illustration}). As Fig
\ref{fig:robustness_performance} shows, our controller successfully transferred to the unseen environments with a low noise level ($\eta_{n}=0.1$), which is similar to that of real RGB-D cameras. Even the in environment with an unrealistic high noise level ($\eta_{n}=0.5$), the reduction in success rate was $19\%$ compared to our baseline ($\eta_{n}=0$), indicating our method has a strong robustness to depth image noises. For the qualitative evaluation of robustness, external disturbances were applied to the needle. Our learned controller was able to grasp the dynamic needle due to perturbation and learn to re-grasp within the task. Qualitative results can be found in our supplementary video.

\subsection{Ablation Study}
\label{sec:ablation_study}

We evaluated the effectiveness of individual components in the ablation study, where ablative baselines are:
\begin{enumerate}
    \item Removal of BC (No BC). In this baseline. we set $\beta_{bc}$ to zero to remove the effect of BC.
    \item Removal of actor gradients (No Actor Grad). We set $\beta_{r}$ to zero to remove the effect of Reinforce algorithm.
    \item Removal of DSA (No DSA). We replaced DSA with the image representation in the DreamerV2 baseline.
    \item Removal of Virtual Clutch (No Virtual Clutch). $H_{clutch}$ was set to zero to remove the effect of Virtual Clutch.
\end{enumerate}

\begin{figure}[!tbp]
  \centering
  \includegraphics[width=0.8\hsize]{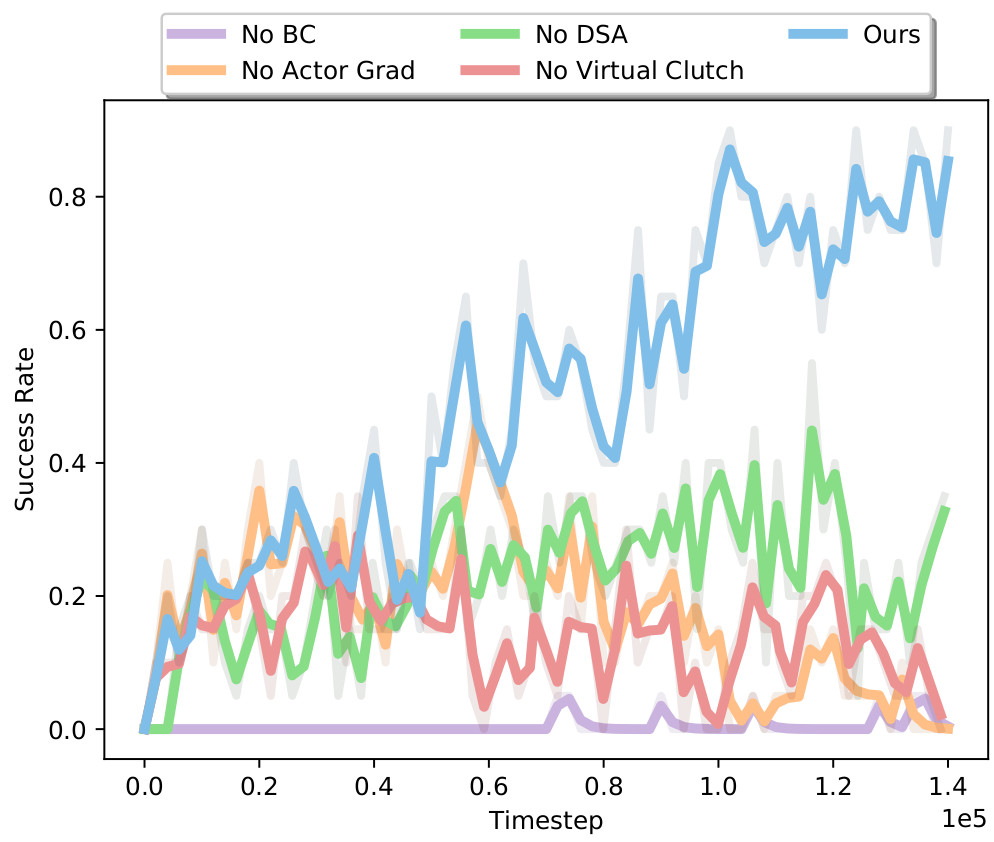}
  \caption{Success rate in the evaluation of our ablation study. Curves are smoothed by Exponential Moving Window with a $0.3$ smoothing factor.}
  \vspace{-0.45cm}
  \label{fig:ablation}
\end{figure}

Fig. \ref{fig:ablation} shows the evaluated success rate during training in our ablation study. The success rate of our method was significantly higher compared to No BC and No Actor Grad, demonstrating the efficient policy learning of our method using the mixed effect of Reinforce Algorithm and BC. The learning of No Virtual Clutch was more unstable compared to ours as its success rate dropped to zero in the course of training. The introduction of Virtual Clutch, on the other hand, effectively eliminates the instability effect due to the significant error between prior and posterior encoded states, resulting in a steady increase in the success rate of the training process. Our method converged faster and performed better than No DSA, indicating that our image representation using DSA was more efficient for the end-to-end learning of visuomotor policy in needle-picking tasks.

\section{Conclusion, Limitation and Discussion}
In this paper, we proposed the first end-to-end learning method of deep visuomotor policy for needle picking. In particular, DreamerfD, which integrated DreamerV2 with demonstrations, was proposed to improve learning efficiency; We proposed DSA to maximally represent control-related visual information in 64x64x3 low-dimensional image space, due to the limited capacity of the world models in DreamerV2; Virtual Clutch was applied to ensure low error between prior and posterior encoded states in DreamerV2. Extensive experiments were carried out to evaluate the performance, in-domain variation adaptation, robustness, and efficiency of individual components of our method. We demonstrated our visuomotor policy can adapt to unseen in-domain variations.

Our method was limited as follows: 1) DSA relied on the segmentation of key objects and more rigorous experiments need to be carried out to evaluate the robustness of segmentation disturbance; when applying DSA to complex scenes in real surgery, data-driven segmentation methods such as MaskRCNN\cite{he2017mask} might yield better performance compared to color-based segmentation; 2) Virtual Clutch was empirically determined by timesteps and incorporating more factors (e.g., the error between prior and posterior encoded states) for our Virtual Clutch might produce better performance. In our future work, we will conduct more rigorous studies on our proposed methods and deploy our method to real robots. Furthermore, we will apply our method to other surgical tasks and general robot manipulation tasks to evaluate its cross-domain generalization.

\section{Acknowledgement}
Thanks for the technical discussion with Adnan Munar and Juan Barragan Noguera on the simulation.

\bibliographystyle{IEEEtran}
\bibliography{Reference}

\appendices
\section{Hyperpxarameters}
\label{sec:appendix_a}
\begin{table}[hbt!]
\scriptsize
\renewcommand{\arraystretch}{1.3}
\label{table:coupling_data_joints_ranges}
\centering
\begin{tabular}{|c| c| c |}
\hline
Name & Symbol & Value \\ 
\hline
\hline
KL Weight & $\beta_{kl}$ & $1$\\\hline
Reinforce Weight & $\beta_{a}$ & $1$\\\hline
Entropy Regularizer Weight & $\beta_{e}$ & $0.002$\\\hline
BC Weight & $\beta_{bc}$ & $1$\\\hline
Virtual Clutch Timestep &$H_{clutch}$& $6$\\\hline
Zoom-in Margin Ratio & $b$ & $0.3$\\\hline
Batch Size &$M$& 70 \\\hline
Batch Length &$N$& 10 \\\hline
Imagine Horizon& $H$& 15\\\hline
Pre-training Steps& $L_{pre}$& 100\\\hline
Gradient Update& $K$& 100\\\hline
Gradient Update Step&$L_{every}$& 50 \\\hline
World Model Learning Rate &$l_{\phi}$& $2\times10^{-4}$\\\hline
Critic Learning Rate &$l_{\xi}$& $4\times10^{-5}$\\\hline
Actor Learning Rate &$l_{\psi}$& $2\times10^{-5}$\\\hline
RSSM Size & - & 512\\\hline

\end{tabular}
\end{table}
\end{document}